\documentclass{bmvc2k}
\title{Five A$^{+}$ Network: You Only Need 9K Parameters for Underwater Image Enhancement}

\addauthor{Jingxia Jiang}{202021114006@jmu.edu.cn}{1\dag}
\addauthor{Tian Ye}{201921114031@jmu.edu.cn}{1\dag}
\addauthor{Jinbin Bai}{jinbin.bai@u.nus.edu}{2\dag}
\addauthor{Sixiang Chen}{201921114013@jmu.edu.cn}{1}
\addauthor{Wenhao Chai}{wenhaochai.19@intl.zju.edu.cn}{3}
\addauthor{Shi Jun}{yunliu@swu.edu.cn}{4}
\addauthor{Yun Liu}{yunliu@swu.edu.cn}{5}
\addauthor{Erkang Chen}{ekchen@jmu.edu.cn}{1,6}
\usepackage{bbding}

\addinstitution{
 School of Ocean Information Engineering,\\
 Jimei University,\\ 
 Xiamen, China
}
\addinstitution{
 Department of Computer Science,\\
 National University of Singapore,\\
 Singapore 
}
\addinstitution{
ZJU-UIUC Institute,\\
Zhejiang University,\\
Hanzhou, China
}
\addinstitution{
the School of Information Science and Engineering, \\
Xinjiang University,\\ 
Uruqmi, China
}
\addinstitution{
 College of Artificial Intelligence, \\
 Southwest University, \\
 Chongqing, China \\
}
\addinstitution{
Fujian Provincial Key Laboratory of Oceanic Information Perception and Intelligent Processing,\\
China
}
\runninghead{~JINGXIA JIANG ET AL.}{9K Parameters for Underwater Image Enhancement}

\begin{document}
\footnotetext{\dag~These authors contributed equally to this work.}

\maketitle

\vspace{-0.5cm}
\begin{abstract}

A lightweight underwater image enhancement network is of great significance for resource-constrained platforms, but balancing model size, computational efficiency, and enhancement performance has proven difficult for previous approaches. In this work, we propose the Five A$^{+}$ Network (FA$^{+}$Net), a highly efficient and lightweight real-time underwater image enhancement network with only $\sim$ \textbf{9k} parameters and $\sim$ \textbf{0.01s} processing time. The FA$^{+}$Net employs a two-stage enhancement structure. The strong prior stage aims to decompose challenging underwater degradations into sub-problems, while the fine-grained stage incorporates multi-branch color enhancement module and pixel attention module to amplify the network's perception of details. To the best of our knowledge, FA$^{+}$Net is the only network with the capability of real-time enhancement of 1080P images. Thorough extensive experiments and comprehensive visual comparison, we show that FA$^{+}$Net outperforms previous approaches by obtaining state-of-the-art performance on multiple datasets while significantly reducing both parameter count and computational complexity. The code is open source at \hyperlink{https://github.com/Owen718/FiveAPlus-Network.}{https://github.com/Owen718/FiveAPlus-Network}.

\end{abstract}

\vspace{-0.4cm}
\section{Introduction}
\label{sec:intro}
Underwater images are often plagued by severe blurring and color distortion, making it difficult to meet the demands of practical applications. With the rise of underwater archaeology~\cite{hariry2021enforcement,yao2022application} and marine ecological research~\cite{todd2019towards,kaiser2011marine,long2011marine}, some researchers have begun to explore how to embed underwater image enhancement algorithms into platforms such as underwater robots. However, due to the limited resources of underwater robots, underwater cameras, and other equipment, traditional deep learning models are challenging to achieve efficient underwater image enhancement on these platforms~\cite{li2019underwater,li2021underwater,huo2021efficient,ye2022underwater}.

One potential solution to this issue is to design lightweight networks with fewer parameters and computations. For example, the Shallow-uwnet~\cite{naik2021shallow} constructed a shallow network by introducing lightweight network components and residual convolution blocks. However, this purely resource-driven approach may not necessarily result in lower computational complexity. Additionally, no specific design has been proposed to target certain degradation phenomena in underwater enhancement tasks, resulting in unsatisfactory visual effects and performance metrics for the restored images.

Constructing a real-time underwater image enhancement framework that simultaneously possesses ultra-lightweight parameters and powerful enhancement ability has been a long-standing challenge in the field.

\begin{figure}
    \centering
   \bmvaHangBox{{\includegraphics[width=10cm]{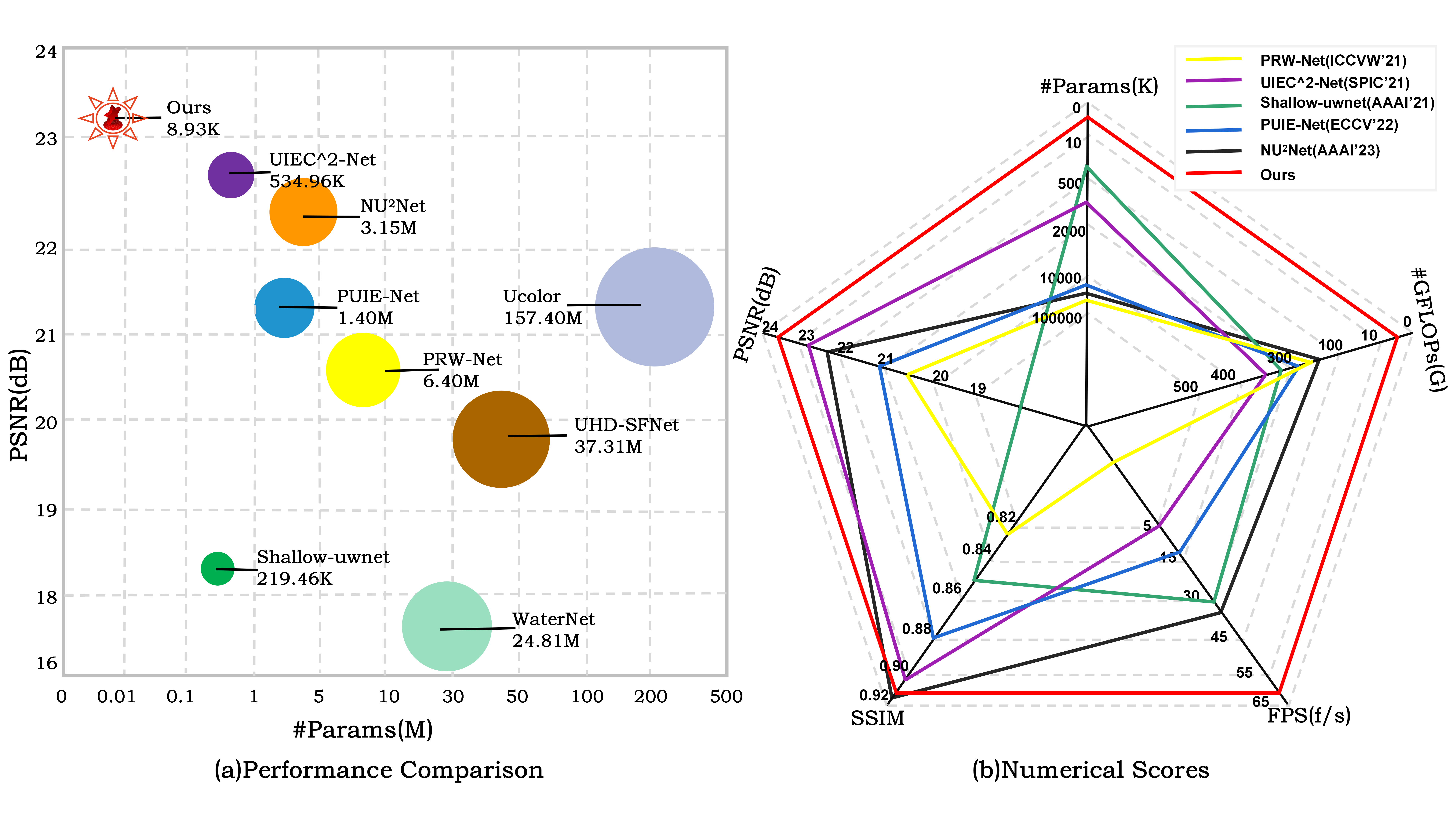}}}
    \caption{\small  Comparison of recent state-of-the-art methods and our method: We report the computational efficiency (${\rm{\# }}Params$, GFLOPs, and FPS) and numerical
scores for two types of restoration quality measurement metrics including PSNR and SSIM, it can be easily observed that our method is remarkably superior to others.}
    \label{BMVCFIG1}\vspace{-0.5cm}
\end{figure}

To overcome this challenge, we decompose the underwater degradations suffered by underwater images into sub-problems based on the characteristics of the Underwater Image Enhancement (UIE) task and design a lightweight and embedded real-time UIE network called the Five A$^{+}$ Network (FA$^{+}$Net). The Five A$^{+}$ symbolizes that our network achieves outstanding performance in terms of PSNR, SSIM, FPS, GFLOPs and Parameters. As demonstrated by Fig.~\ref{BMVCFIG1}, FA$^{+}$Net achieves state-of-the-art performance while significantly reducing both parameter count and computational complexity by an order of magnitude compared to previous methods by 10-100 $\times$, with a total parameter count of less than 9K.

To reduce model complexity, some computationally expensive operators and operations such as large-kernel convolutions~\cite{ding2022scaling,feng2022lkasr,guo2022visual} and self-attention~\cite{shaw2018self,zhang2019self,vaswani2021scaling,10095605,10096828,chen2022snowformer} were discarded while channel dimensions are precisely restricted to control parameter count.  For UIE, the problem can be addressed by breaking it down into sub-problems, effectively solving the mixed degradation problem by separately correcting color distortion and restoring details of the degraded image. For the strong prior stage, two complementary components are proposed: Multi-Branch Color Enhancement Module(MCEM)  and Multi-scale Pyramid Module (MPM). MCEM is an effective module for serious color distortion of underwater images. and MPM enables processing of input feature maps at multiple scales, thus capturing detail information at varying scales to enhance the model's perception of image details. In this way, our strong prior-based designs endow the network with highly effective restoring capabilities for underwater degradations.

Recently, some researchers have proposed the separation of global background light and texture in the Fourier domain~\cite{li2023embedding,suvorov2022resolution}. Specifically, global background light is represented by amplitude, while texture is intertwined with phase. By separating them in the Fourier domain, Gaussian noise can be avoided when enhancing color, whilst providing abundant global information.
To better extract valuable feature information from different component outputs and capture global contextual information, we design a Spatial-frequency Domain Feature Interaction Module(SDFIM). We utilize adjustable hyperparameter $\alpha$ to control the fusion of spatial domain and frequency domain information. Additionally, Fast Fourier Convolution (FFC)~\cite{chi2020fast} is adopted to enlarge the receptive field of our network to entire resolution,  significantly amplifying the network's perception ability. Notably, FFC's inductive bias elevates the network's generalization performance, requiring less training data and computation.
Although these operations have shown basic success in addressing the mixed degradation issue, the complex underwater environment is often impacted by multiple factors causing some challenging detail problems. In these cases, traditional single-stage networks may struggle to accurately capture tiny objects, intricate colors, and textures. To further enhance the model's performance, we introduce a fine-grained stage for more in-depth image analysis, aiming to better manage these intricate detail issues. In the fine-grained stage, MCEM and Pixel Attention~\cite{qin2020ffa} mechanism are incorporated to assist the model in comprehending each image element and detail more effectively, thereby improving the model's performance and generalization capability. By introducing the fine-grained stage, substantial progress is made in the model's ability to tackle complex underwater images, as it is better equipped for handling intricate detail issues.

The integration of these two stages not only present novel design ideas for underwater image enhancement but also expand the horizon of potential research in this field.
Notably, 8.9K parameters makes FA$^{+}$Net could be embeddable into edge devices, and we are the only framework capable of real-time enhancement of 1080P-sized images on RTX 3090. Our model also possesses high throughput, enabling faster inference and processing of input data, thereby meeting the requirements of mobile platforms such as underwater shooting devices and robot platforms.

This paper presents the following key contributions:
\begin{itemize}
    \item We introduce FA$^{+}$Net, that compresses the parameter count of an enhancement model to 8.9K, which is on the order of magnitude of $10-100\times$ less than previous methods.
    
    \item We propose a two-stage architecture that provides novel designs and directions for underwater image enhancement. The strong prior stage decomposes mixed degradation into sub-problems, while the fine-grained stage seeks to improve the network’s perception of details.

    \item Our network, FA$^{+}$Net, is the only one that can perform real-time enhancement of 1080P images on RTX 3090, achieving better results on multiple datasets with fewer parameters and computations, thus being suitable for deployment on mobile platforms.
\end{itemize}

\vspace{-0.5cm}
\section{Related Work}
\subsection{Learning-based Underwater Image Enhancement Method}
\vspace{-0.1cm}
With the successful application of deep learning in high-level computer vision tasks~\cite{zou2023object}, an increasing number of researchers have begun to apply it to low-level computer vision tasks~\cite{jin2022estimating,jin2022unsupervised,jin2022shadowdiffusion,jin2023structure,Ye_2022_ACCV,10.1007/978-3-031-19800-7_8}, such as underwater image enhancement~\cite{li2020underwater,guo2023uranker,jiang2023rsfdm,ye2022underwater}. For instance, Li et al.~\cite{li2019underwater} proposed the WaterNet model, which used adaptive filters and deep convolutional neural networks to improve image quality and reduce noise. Jiang et al.~\cite{jiang2022two} designed a novel domain adaptation framework based on transfer learning to transform aerial image deblurring into realistic underwater image enhancement. Despite their varying degrees of success in terms of performance metrics, these approaches fail to incorporate dedicated modules for addressing color shift and texture loss of degraded images. Li et al.~\cite{li2021underwater}presented an underwater image enhancement network via medium transmission-guided multi-color space embedding, named Ucolor. Huo et al.~\cite{huo2021efficient} employed wavelet-enhanced learning units to decompose hierarchical features into high-frequency and low-frequency components, and then strengthen them with normalization and attention mechanisms. Although this approach has shown excellent visual effects, its extensive network parameters (6.30M) and computational requirements (223.37G) make it unsuitable for existing underwater devices. Moreover, it cannot effectively address the issue of color distortion.

\vspace{-0.38cm}
\subsection{Efficient Neural Network For Image Restoration}\vspace{-0.1cm}
Efficient neural network for image restoration~\cite{Ye_2022_ACCV,cui2022illumination,wei2022uhd} is a recent development in deep learning-based image restoration and has been demonstrated to achieve state-of-the-art performance while requiring fewer computational resources than other methods. For example, Song et al.~\cite{song2020efficient} proposed an efficient residual dense block search algorithm with multiple objectives to identify fast, lightweight, and accurate networks for image super-resolution. Guo et al.~\cite{7782813} presented an effective low-light image enhancement method (LIME) that estimated the illumination of each pixel individually and refined it using a structure prior. Naik et al.~\cite{naik2021shallow} proposed a lightweight underwater enhancement framework by introducing lightweight components and residual blocks.

What's more, FA$^{+}$Net has only 8.9K parameters, making it even more competitive than its counterparts in terms of training speed and computational resource consumption, thus offering potential applications in embedded equipment or mobile applications.

\vspace{-0.3cm}
\subsection{Fast Fourier Convolution}
\vspace{-0.1cm}
In order to address the low efficacy in connecting two distant locations in the network. Chi et al.~\cite{chi2020fast} proposed a novel convolutional operator dubbed as Fast Fourier Convolution (FFC), which has the characteristics of non-local receptive fields and cross-scale fusion within the convolutional unit. Furthermore, modern image inpainting systems commonly struggle with large missing areas, complex geometric structures, and high-resolution images. To alleviate this issue,  Suvorov et al.~\cite{suvorov2022resolution} proposed a new method termed large mask inpainting that is based on a new inpainting network architecture relying on FFCs. When dealing with the challenging task of joint luminance enhancement and noise removal whilst remaining efficient. Li et al.~\cite{li2023embedding} devised a new solution, UHDFour, which differs from existing approaches that take a spatial domain-oriented approach. Specifically, UHDFour is motivated by a few unique characteristics of the Fourier domain, such as the fact that most luminance information is concentrated in amplitudes while noise is closely related to phases.

\begin{figure}
    \centering
   \bmvaHangBox{{\includegraphics[width=12.8cm]{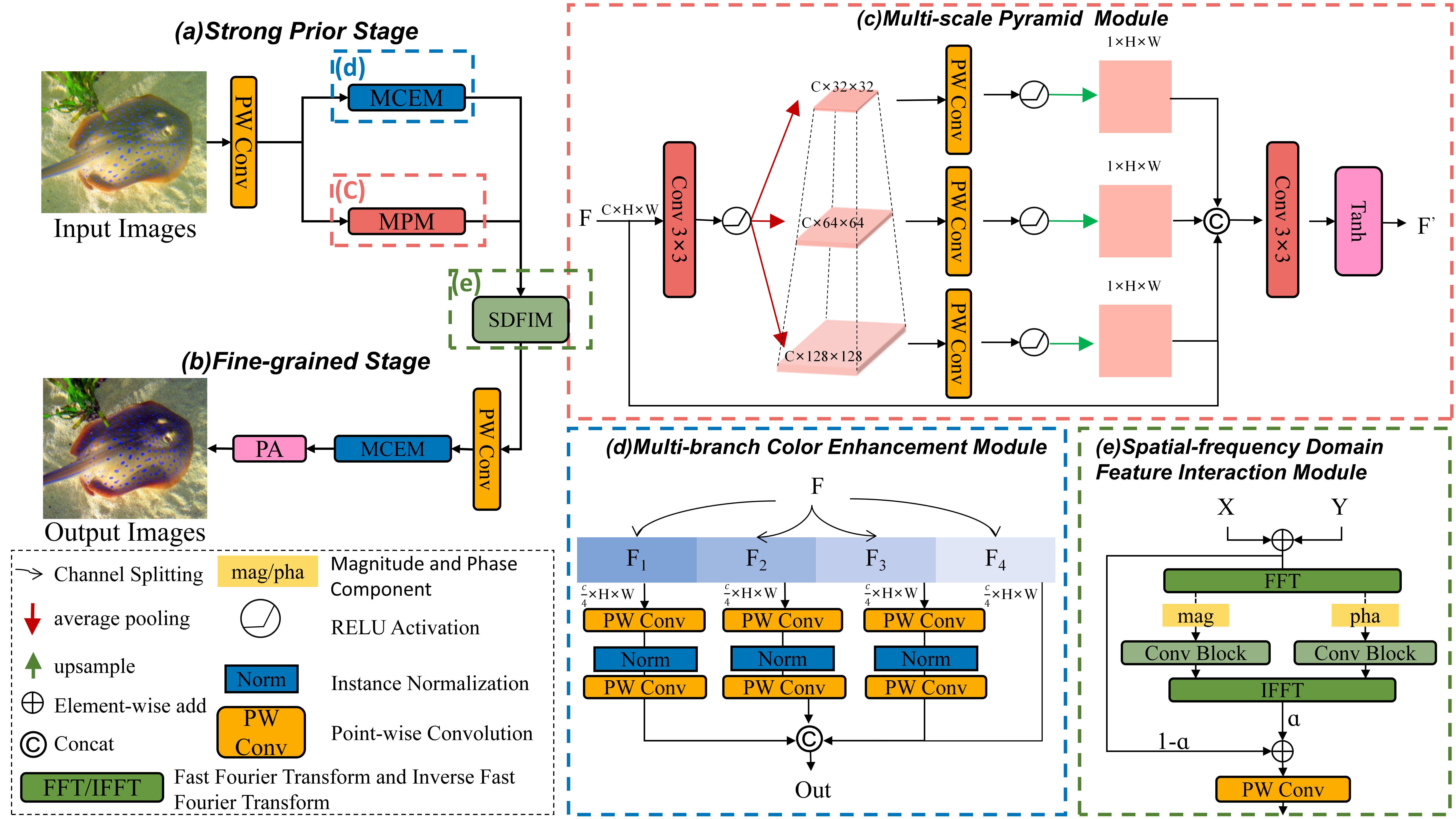}}}
    \caption{\small The overall architecture of Five A$^{+}$ Network: FA$^{+}$Net is composed of a strong prior stage and a fine-grained stage, augmented by the efficient Spatial-frequency Domain Interaction Module. The core components of the network comprise: (c) MPM captures granular details across various scales, endowing the model with potent detail perception; (d) MCEM perform consecutive processing of individual image pixels, thereby enabling our network to achieve precise color restoration; and (e) SDFIM aids the network in sifting valuable feature information from the outputs of diverse components and acquiring global contextual features, and $\alpha$ is a hyperparameter that controls the fusion ratio of spatial-frequency domain information.}
    \label{BMVCFIG2}\vspace{-0.5cm}
\end{figure}

\vspace{-0.1cm}
\section{FA$^{+}$Net: An Ultra-lightweight Real-time Enhancement Network}
\subsection{Motivation}
The limited computing resources of embedded platforms, such as underwater robots, pose a significant challenge in achieving high-quality image enhancement using traditional deep learning models. Consequently, the development of lightweight yet powerful models has been prompted by recent methods~\cite{cui2022illumination,song2020efficient,7782813}. As a notable contribution in this regard, FA$^{+}$Net has been demonstrated to be an efficient and innovative solution, as depicted in Fig.~\ref{BMVCFIG2}.

To ensure the computational efficiency of our model, we first remove several computationally expensive operators and operations, such as large kernel convolutions and self-attention mechanisms, and further imposed constraints on the channel dimension for accurate parameter control. Furthermore, to address the mixed degradation problem effectively, we adopt a divide-and-conquer strategy to separately enhance color and restore details from degraded images. Moreover, to improve the performance of our model, we propose a fine-grained stage for comprehensive image analysis. Collectively, our approach allows for effective color enhancement and detail restoration, even under extreme underwater conditions.
\vspace{-0.3cm}
\subsection{Model Structure}
\subsubsection{Multi-Scale Pyramid Module}
\vspace{-0.1cm}
To recover fine details in degraded underwater images, we propose a Multi-scale Pyramid Module(MPM) in the strong prior stage. By downsampling the input image to different sizes, the network can capture features at multiple scales and resolutions, which is critical for improving the appearance of objects with different sizes and shapes in challenging underwater scenarios. To achieve real-time performance, we designed the MPM as a three-branch structure with down-sampled target size of $32\times32$, $64\times64$, and $128\times128$. The selection of this structure is based on a series of careful ablation experiments reported in~\ref{MPDC}, which ensured a good trade-off between performance and effectiveness. Further information can be found in the supplementary material.

\vspace{-0.2cm}
\subsubsection{Multi-branch Color Enhancement Module}
\vspace{-0.1cm}
The attenuation rates of different wavelengths of light in underwater environments vary, with red light experiencing the fastest attenuation and blue and green light experiencing the slowest~\cite{jiang2022underwater}. This results in conspicuous differences in the R, G, and B channels, leading to poor contrast and color distortion in underwater images, which has been a largely unaddressed issue in previous methods~\cite{ancuti2017color,8296508,9001231,li2021underwater}. 

To overcome this limitation, we propose the MCEM, which employs a branch enhancement strategy to better capture the color feature distribution across R, G, and B channels. In the MCEM, we use $1\times1$ convolutions to perform consecutive processing of individual image pixels, enabling the capture of color information at each pixel. This approach is similar to the underlying operation of a multi-layer perception~\cite{taud2018multilayer}, allowing our network to achieve accurate color reproduction, which is particularly crucial for color-sensitive underwater image enhancement tasks. Additionally, we opt not to use $3\times3$ convolutions due to their increased parameter burden. The MCEM partitions the input into four branches, as shown in Fig.~\ref{BMVCFIG2}(d), with weights not shared between each branch. The effectiveness of this module is demonstrated in the supplementary material.

\vspace{-0.2cm}
\subsubsection{Spatial-frequency Domain  Feature Interaction Module}
\vspace{-0.1cm}
Recent studies have shown that global background lighting and textures in underwater images can be partially decomposed in the Fourier domain, as evidenced by recent works such as \cite{chi2020fast,li2023embedding,suvorov2022resolution,zhou2022adaptively,huang2022deep}. However, current methods for restoring degraded images mostly rely on spatial domain processing, and traditional convolutional approaches tend to overlook the rich global information present in the Fourier domain. To address this issue, we propose a cross-domain design approach called Spatial-Frequency Domain Interaction Module(SDFIM) for underwater image enhancement. By fusing feature information in the Fourier domain, SDFIM achieves receptive field coverage of the entire image, which improves the network's perceptual quality and parameter efficiency. The hyperparameter $\alpha$ in SDFIM controls the fusion ratio of spatial-frequency domain information, and its varying values generate different visual effects. Furthermore, the induction bias of Feature Fusion Convolution (FFC) enhances the network's generalization performance, thereby reducing the requirements for extensive training data and computation.

The key operations of SDFIM are as follows, given the feature $X \in {R^{C {\rm{ \times }}H{\rm{ \times }}W}}$ and $Y \in {R^{C{\rm{ \times }}H{\rm{ \times }}W}}$:
\begin{equation}
F' = X + Y
\end{equation}
\begin{equation}
{F_{MAG}},{F_{PHA}} = {{\rm{f}}_{FFT}}(F')
\end{equation}
\begin{equation}
{F_{OUT}} = \alpha[{f_{IFFT}}({f_{FC}}({F_{MAG}}),{f_{FC}}({F_{PHA}}))] + (1 - \alpha)F'
\end{equation}
where $F_{MAG}$ and $F_{PHA}$ represent the magnitude component and phase component of the feature, respectively. ${\rm{f}}_{FFT}(\cdot)$ denotes the fast Fourier transform, ${f_{FC}(\cdot)}$ represents the Fourier domain convolution operation, and ${f_{IFFT}(\cdot)}$ denotes the inverse fast Fourier transform. The hyperparameter $\alpha$ controls the fusion ratio of spatial-frequency domain information.
\vspace{-0.2cm}
\section{Experiments}
\vspace{-0.1cm}
\subsection{Experiment details}\label{subsub1}
All experiments are implemented using the PyTorch \cite{NEURIPS2019_bdbca288} framework with a single NVIDIA GTX A100GPU (40GB). During training, the training epochs are set to 400, and the total batch size is 72. We use Adam optimizer as the optimization algorithm. The learning rate is set to $4\times 10^{-4}$ at first, and the default values of $\beta_1$ and $\beta_2$ are 0.5 and 0.999, respectively. We used CyclicLR to adjust the learning rate, with an initial momentum of 0.9 and 0.999. Data augmentation included horizontal flipping, and randomly rotating the image to $90, 180$, and $270$ degrees.

During the training process, the input data was randomly cropped as $256\times 256$ patches from original images. UIEB Datasets~\cite{li2019underwater} contains 890 high-resolution raw underwater images and corresponding high-quality reference images, and 60 challenge images (C60) for which no corresponding reference images were obtained. Li et al. carefully selected 45 authentic underwater images, named U45~\cite{li2019fusion}. It is partitioned into three subsets according to the color cast of underwater degradation, low contrast, and blur effects: green, blue, and haze. Then, 800 pairs of original images and clear images were extracted from UIEB to train the model. The remaining 90 images in UIEB named T90 were used to test the effect of our method on degraded images. In order to evaluate the generalization performance of FA$^{+}$Net, we used the C60 and U45 datasets for testing.
\vspace{-0.2cm}
\begin{figure}
    \centering
   \bmvaHangBox{{\includegraphics[width=12.8cm]{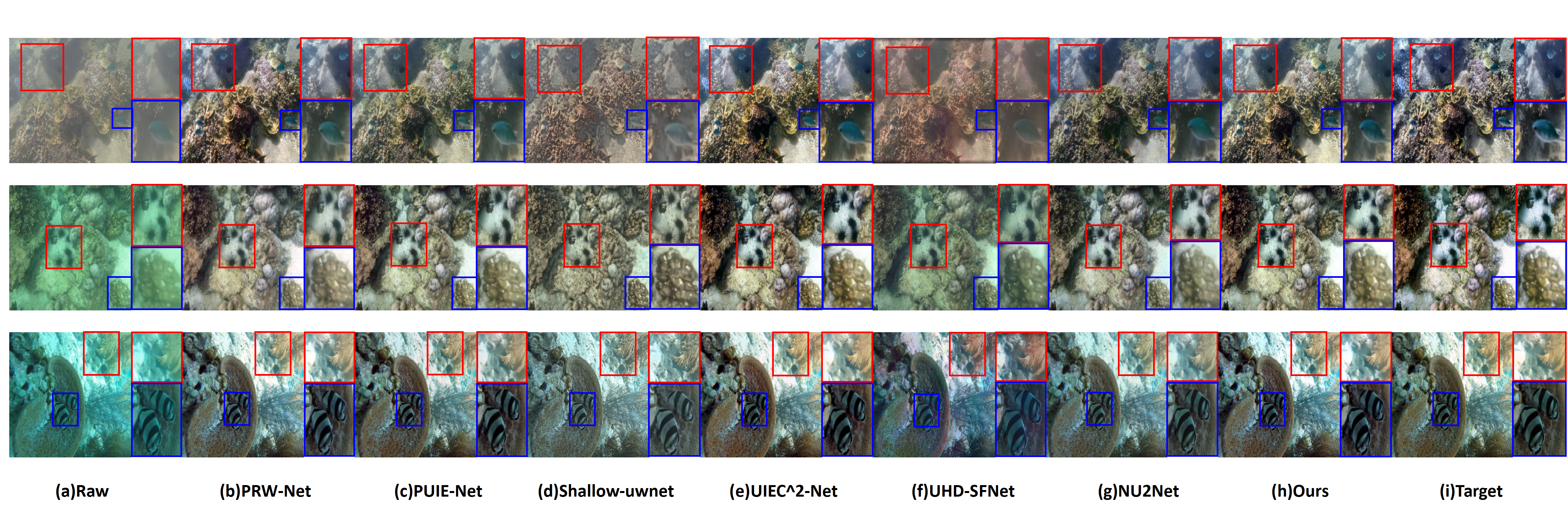}}}
    \caption{\small Visual comparison of UIE networks on T90.}\vspace{-0.45cm}
    \label{BMVCFIG3}
\end{figure}
\subsection{Evaluation metrics}\label{subsub2}
In order to acquire quantitative measurements, we use Peak Signal-to-Noise Ratio (PSNR) \cite{korhonen2012peak}, Structural Similarity Index (SSIM) \cite{wang2004image}, the Mean Squared Error (MSE) \cite{4308985}, Underwater Color Image Quality Evaluation (UCIQE) \cite{7300447}, and Underwater Image Quality Metric (UIQM) \cite{7305804} as performance metrics for image quality. PSNR is a full-reference image quality evaluation metric based on errors between corresponding pixels. The higher the PSNR score, the better the image quality. SSIM measures the visual quality of three features of an image: brightness, contrast, and structure. A higher SSIM value indicates a higher similarity between the enhanced and reference images. UCIQE mainly measures the degree of detail and color recovery of distorted images. UCIQE is one of the most comprehensive image evaluation standards. UIQM is required to evaluate color, sharpness, and contrast.

\vspace{-0.3cm}
\subsection{Comparison with SOTA methods}\label{subsub3}
We compared FA$^{+}$Net with several state-of-the-art methods, including traditional methods and deep learning methods. Traditional methods included UDCP \cite{drews2013transmission}, IBLA \cite{peng2017underwater}, SMBL \cite{song2020enhancement} and MLLE \cite{zhang2022underwater}, and deep learning methods included UWCNN \cite{li2020underwater}, Water-Net \cite{li2019underwater}, PRW-Net \cite{huo2021efficient}, Shallow-uwnet \cite{naik2021shallow}, Ucolor \cite{li2021underwater}, UIEC\textasciicircum{}2-Net \cite{wang2021uiec}, UHD-SFNet~\cite{wei2022uhd}, PUIE-Net \cite{fu2022uncertainty} and the latest NU2Net~\cite{guo2023uranker} for underwater image enhancement. We present the objective metrics comparison with previous SOTA methods in Table~\ref{Table1}. From that, we can observe that our method achieves the best results on PSNR, MSE and UCIQE metrics, proving that the proposed architecture has good results with detailed textures, restoring promising contrast and color of images. Compared with the last method NU2Net on T90, we exceed 0.642dB, 0.01 and 0.029 on PSNR, MSE and UCIQE respectively. 
\begin{table}[]
  \centering
\caption{\small{Experimental results on T90~\cite{li2019underwater}, C60~\cite{li2019underwater}, and U45~\cite{li2019fusion} datasets, best and second best results are marked in \textcolor{red}{red} and \textcolor{blue}{blue} respectively. $\uparrow$ represents the higher is the better as well as $\downarrow$ represents the lower is the better. The efficiency evaluation uses 720P images as input on RTX 3090.}}\label{Table1}
\resizebox{12.1cm}{!}{
\begin{tabular}{l|ccccc|cc|cc|cccc}
\hline
\multicolumn{1}{c|}{}                                   & \multicolumn{5}{c|}{\textbf{T90}}                                          & \multicolumn{2}{c|}{\textbf{C60}} & \multicolumn{2}{c|}{\textbf{U45}} & \multicolumn{4}{c}{\textbf{Efficiency}}      \\ \cline{2-14} 

\multicolumn{1}{l|}{{\textbf{Methods}}} & PSNR$\uparrow$           & SSIM$\uparrow$           & MSE $\downarrow$           & UCIQE↑                        & UIQM$\uparrow$           & UCIQE$\uparrow$                       & UIQM$\uparrow$                        & UCIQE↑                      & UIQM↑                       & GFLOPs(G) $\downarrow$     & \#Params(M) $\downarrow$    & \#Runtime(s) $\downarrow$   & FPS(f/s)$\uparrow$       \\ \hline
UDCP(ICCVW'13)\cite{drews2013transmission}                                                                  & 13.415          & 0.749          & 0.228          & 0.572                         & 2.755          & 0.560                       & 1.859                       & 0.574                       & 2.275                       & -   & -               & 42.13s          & -               \\
IBLA(TIP'17)\cite{peng2017underwater}                                                                    & 18.054          & 0.808          & 0.142          & 0.582                         & 2.557          & {\textcolor{blue}{0.584}}          & 1.662                       & 0.565                       & 2.387                       & -              & -               & -               & -               \\
WaterNet(TIP'19)\cite{li2019underwater}                                                                & 16.305          & 0.797          & 0.161          & 0.564                         & 2.916          & 0.550                       & 2.113                       & 0.576                       & 2.957                       & 193.70G        & 24.81M          & 0.680s          & -               \\
SMBL(TB'20)\cite{song2020enhancement}                                                                     & 16.681          & 0.801          & 0.158          & 0.589                         & 2.598          & 0.571                       & 1.643                       & 0.571                       & 2.387                       & -              & -               & -               & -               \\
UWCNN(PR'20)\cite{li2020underwater}                                                                    & 17.949          & 0.847          & 0.221          & 0.517                         & 3.011          & 0.492                       & 2.222                       & 0.527                       & 3.063                       & -              & -               & -               & -               \\
PRW-Net(ICCVW'21)\cite{huo2021efficient}                                                               & 20.787          & 0.823          & 0.099          & 0.603                         & {\textcolor{red}{3.062}} & 0.572                       & {\textcolor{red}{2.717}}              & 0.625                       & 3.026                       & 223.4G         & 6.30M           & 0.216s          & 4.624           \\
Shallow-uwnet(AAAI'21)\cite{naik2021shallow}                                                          & 18.278          & 0.855          & 0.131          & 0.544                         & 2.942          & 0.521                       & 2.212                       & 0.545                       & 3.109                       & 304.75G        & {\textcolor{blue}{0.22M}}           & 0.031s          & 31.836          \\
Ucolor(TIP'21)\cite{li2021underwater}                                                                  & 21.093          & 0.872          & 0.096          & 0.555                         & {\textcolor{blue}{3.049}} & 0.530                       & 2.167                       & 0.554                       & 3.148                       & 443.85G        & 157.42M         & 2.758s          & -               \\
UIEC\textasciicircum{}2-Net(SPIC'21)\cite{wang2021uiec}                                            &{\textcolor{blue}{22.958}} & 0.907          & {\textcolor{blue}{0.078}} & {\textcolor{blue}{0.599}}                & 2.999          & 0.580                       & {\textcolor{blue}{2.228}}              &  {\textcolor{blue}{0.604}}              & 3.125                       & 367.53G        & 0.53M           & 0.174s          & 5.742           \\
MLLE(TIP'22)\cite{zhang2022underwater}                                                                    & 19.561          & 0.845          & 0.115          & 0.592 & 2.624          & 0.581                       & 1.977                       & 0.597                       & 2.454                       & -              & -               & -               & -               \\
UHD-SFNet(ACCV'22)\cite{wei2022uhd}                                                              & 18.877          & 0.810          & 0.144          & 0.559                         & 2.551          & 0.528                       & 1.741                       & 0.585                       & 2.826                       & {\textcolor{blue}{15.24G}}         & 37.31M          & 0.059s          & 16.769          \\
PUIE-Net(ECCV'22)\cite{fu2022uncertainty}                                                               & 21.382          & 0.882          & 0.093          & 0.566                         & 3.021          & 0.543                       & 2.155                       & 0.563                       & {\textcolor{red}{3.192} }           & 423.05G        & 1.40M           & 0.071s          & 14.194          \\
NU2Net(AAAI'23,Oral)\cite{guo2023uranker}                                                            & 22.419          & {\textcolor{red}{0.923}} & 0.086          & 0.587                         & 2.936          & 0.555                       & 2.222                       & 0.593                       & {\textcolor{blue}{3.185}}              & 146.64G        & 3.15M           & {\textcolor{blue}{0.024s}}          & {\textcolor{blue}{42.345}}          \\ \hline
Ours                                                                            &{ \textcolor{red}{23.061}} & {\textcolor{blue}{0.911}} & {\textcolor{red}{0.076}} & {\textcolor{red}{0.616}}               & 2.828          & {\textcolor{red}{0.593}}              & 2.088                       & {\textcolor{red}{0.609}}              & 3.174                       & {\textcolor{red}{8.33G}} & {\textcolor{red}{0.009M}} & {\textcolor{red}{0.016s}} & {\textcolor{red}{60.724}} \\ \hline
\end{tabular}}
\vspace{-0.3cm}
\end{table}

\begin{table}[!]
\centering
\caption{\small{Performance comparison is tested on RTX 3090 using 1080P resolution (1920$\times$1080) images, best and second best results are marked in red and blue respectively.}}\label{Table4}
\resizebox{8cm}{!}{
\begin{tabular}{l|cccc}
\hline
\multicolumn{1}{c|}{}                                   & \multicolumn{4}{c}{For 1080P Real-time Test}                                                                                                                                                           \\ \cline{2-5} 
\multicolumn{1}{l|} {Methods} & GFLOPs(G)$\downarrow$ & \#Params(M)$\downarrow$ &  \#Runtime(s)$\downarrow$ &FPS(f/s)$\uparrow$ \\ \hline
Shallow-uwnet(AAAI'21)\cite{naik2021shallow}                                                                                 & 685.70G                          &  {\textcolor{blue}{0.22M}}                                & 0.0745s                              & 13.4188                           \\
UHD-SFNet(ACCV'22)\cite{wei2022uhd}                                                                                     &{\textcolor{red}{15.42G}}                           & 37.31M                               & 0.0666s                               &  15.0045                           \\
NU2Net(AAAI'23,Oral)\cite{guo2023uranker}                                                                                   & 29.95G                            & 3.15M                                & {\textcolor{blue}{0.0516s}}                               & {\textcolor{blue}{19.3495}}                           \\ \hline
Ours                                                                                                   & {\textcolor{blue}{18.74G}}                    & {\textcolor{red}{0.009M}}                     & {\textcolor{red}{0.0333s}}                      & {\textcolor{red}{29.9431}}                  \\ \hline
\end{tabular}}
\vspace{-0.5cm}

\end{table}

According to the data in Table~\ref{Table1}, our proposed FA$^{+}$Net outperforms all other designs in terms of efficiency. The amount of parameters possessed by FA$^{+}$Net is only \textbf{1/17500} of that of Ucolor, yet manifests a considerable qualitative improvement. In comparison with Shallow-uwnet, the quantity of parameters has reduced by more than \textbf{1/20}, but the PSNR index is 4.783dB higher, which clearly shows the superiority and viability of our method. What's even more noteworthy is that from Table~\ref{Table4}, we can observe that FA$^{+}$Net is the \textbf{sole} network able to carry out real-time enhancement for 1080P size images.

Additionally, we also gave an intuitive comparison with previous SOTA methods in terms of visual effects. As seen in Fig.~\ref{BMVCFIG3}, Shallow-net was incapable of sufficiently restoring underwater images due to its straightforward network structure; the intensified image saturation was low, and the edge processing effect was subpar; the enhancement result presented by PRW-Net appeared layered; the image processed by UHD-SFNet still contains some local patches; PRW-Net and PUIE-Net demonstrated poor perception of details, resulting in significant erosion of texture details in the augmented photographs. On the other hand, NU2Net lacked precise color control, hence leading to visible chromatic deviations that can be noticed by the human eye. Our method exhibited quite compatible color and detail recovery, enhancing the entire degraded image, and making its contrast and texture details meet the sensory requirements of the human eye. That is credited to our carefully designed MCEM, MPM, and SDFIM. More visual comparisons are available in the supplementary material.

\vspace{-0.4cm}
\subsection{Ablation Study}
\subsubsection{Effectiveness of downsampling size in MPM}\label{MPDC}\vspace{-0.1cm}
In order to explore the influence of down-sampling size on MPM performance, we adjusted the dimensions and examined their impact. As shown in Table~\ref{Table2},  the effectiveness of the erosion model increases with a growing down-sampling size, though at the cost of increased computation time. Once the down-sampling size reached 256, the model's performance gradually shifts toward a plateau. Taking into account efficiency and performance, we eventually had set the multi-scale feature pyramid at the $32\times32$, $64\times64$, and $128\times128$ size.

\begin{table}[]
\centering
\caption{\small{Ablation study on different value of $\alpha$. The efficiency of the models is measured on RTX A100GPU, \Checkmark means that the feature size is selected for the Multi-scale Pyramid Module.}}\label{Table2}
\resizebox{12.5cm}{!}{
\begin{tabular}{l|cccccccc|cc|cc}
\hline
Model & \multicolumn{8}{c|}{Multi-scale Pyramid Configuration} & \multicolumn{2}{c|}{Metrics} & \multicolumn{2}{c}{Efficiency} \\ \cline{2-13} 
                       & 2  & 4 & 8 & 16 & 32 & 64 & 128 & 256 & PSNR$\uparrow$        & SSIM$\uparrow$        & \#Runtime(s)$\downarrow$   & FPS(f/s)$\uparrow$  \\ \hline
a)                     & \Checkmark  & \Checkmark & \Checkmark &    &    &    &     &     & 22.766        & 0.906        & 0.0132s          & 75.398      \\
b)                     &    & \Checkmark & \Checkmark& \Checkmark  &    &    &     &     & 22.811        & 0.908        & 0.0137s          & 72.941      \\
c)                     &    &   & \Checkmark & \Checkmark  & \Checkmark  &    &     &     & 22.937        & 0.908        & 0.0146s          & 68.167      \\
d)                     &    &   &   & \Checkmark  &\Checkmark  & \Checkmark  &     &     & 22.896        & 0.910        & 0.0158s          & 62.952      \\
e)                     &    &   &   &    &    & \Checkmark  &\Checkmark   & \Checkmark  & 23.001        & 0.912        & 0.0321s          & 31.060      \\ \hline
Ours                   &    &   &   &    & \Checkmark  & \Checkmark  & \Checkmark   &     & 23.061        & 0.911        & 0.0251s          & 39.751      \\ \hline
\end{tabular}}
\end{table}
\vspace{-0.4cm}
\subsubsection{Effectiveness of the value of $\alpha$
in SDFIM}\vspace{-0.1cm}
Our SDFIM employs adjustable hyperparameters $\alpha$ to regulate the fusion of spatial domain and frequency domain information. Table~\ref{Table3} illustrates the effect of distinct $\alpha$ values on model performance. Notably, the model attains the best performance at $\alpha$ of 0.4. Different values of $\alpha$ yield divergent outcomes for the network.

\vspace{-0.4cm}
\subsubsection{Limitation}
Although FA$^{+}$Net has exhibited its effectiveness and exceptional performance in underwater image enhancement tasks through experiments on multiple datasets, it is still restricted owing to insufficient training dataset size and unrefined model optimization. Specifically, FA$^{+}$Net may require more model design and optimization to improve its performance in handling complex underwater image detail problems, such as those containing small objects, complex colors, and textures. Additionally, even though FA$^{+}$Net exhibits efficiency and flexibility on resource-constrained mobile platforms, further experiments should be conducted to validate its performance and dependability in practical applications. 

In the future, we may consider appending more adaptive settings to transform FA$^{+}$Net into a universal enhancement framework, thus enhancing its applicability and scalability.

\begin{table}[]
\centering
\caption{\small{Ablation study on the different value of $\alpha$. Underline indicates the best result.}}\label{Table3}
\resizebox{12cm}{!}{
\begin{tabular}{c|cccccccccc}
\hline
{\textbf{$\alpha$}}                            & 0      & 0.1    & 0.2    & 0.3    & 0.4    & 0.5    & 0.6    & 0.7    & 0.8    & 0.9    \\ \hline
\textbf{PSNR} & 22.831 & 23.051 & 22.900 & 22.938 & \underline{23.061}& 22.810 & 0.909  & 22.812 & 22.842 & 22.936 \\
\textbf{SSIM} & 0.908  & 0.910  & 0.907  & 0.908  & \underline{0.911}  & 0.904  & 22.985 & 0.906  & 0.907  & 0.909  \\ \hline
\end{tabular}}
\end{table}
\vspace{-0.5cm}
\section{Conclusion}
\vspace{-0.2cm}
In this paper, we present a groundbreaking underwater real-time enhancement framework, FA$^{+}$Net, designed to surpass previous methodological constraints. Our approach utilizes a divide-and-conquer strategy to effectively address mixed degradation issues by enhancing the color and restoring details of degraded images separately. Notably, FA$^{+}$Net stands as the solitary network proficient in real-time enhancement of 1080P images. Furthermore, our proposed model contains less than 9K parameters while exceeding state-of-the-art methods with fewer computations and parameters. We believe that the FA$^{+}$Net provides new design ideas and directions for underwater image enhancement and will significantly benefit practical applications in underwater archaeology and marine ecological research.

\bibliography{FIVE_APLUS}
\end{document}